\documentclass{article}

\usepackage[preprint]{neurips_2024}

\usepackage[numbers]{natbib}

\usepackage[utf8]{inputenc} % allow utf-8 input
\usepackage[T1]{fontenc}    % use 8-bit T1 fonts
\usepackage{hyperref}       % hyperlinks
\usepackage{url}            % simple URL typesetting
\usepackage{booktabs}       % professional-quality tables
\usepackage{amsfonts}       % blackboard math symbols
\usepackage{nicefrac}       % compact symbols for 1/2, etc.
\usepackage{microtype}      % microtypography
\usepackage{xcolor}         % colors

\usepackage{mathtools}
\usepackage{enumitem}
\usepackage{graphicx}
\usepackage{amsmath}
\usepackage{algorithm}
\usepackage{algorithmic}
\usepackage{multirow}
\usepackage{adjustbox}
\usepackage[position=top,captionskip=-3pt, textfont=footnotesize]{subfig}

\title{Robust Ante-hoc Graph Explainer \\using Bilevel Optimization}

\author{%
  Kha-Dinh Luong \\
  Department of Computer Science\\
  University of California, Santa Barbara\\
  \And
  Mert Kosan \\
  Department of Computer Science\\
  University of California, Santa Barbara\\
  \AND
  Arlei Lopes Da Silva \\
  Department of Computer Science \\
  Rice University, Houston \\
  \And
  Ambuj Singh \\
  Department of Computer Science\\
  University of California, Santa Barbara\\
}

\DeclareMathOperator*{\argmin}{arg\,min}

\begin{document}

\maketitle

\begin{abstract}
  Explaining the decisions made by machine learning models for high-stakes applications is critical for increasing transparency and guiding improvements to these decisions. This is particularly true in the case of models for graphs, where decisions often depend on complex patterns combining rich structural and attribute data. While recent work has focused on designing so-called post-hoc explainers, the broader question of what constitutes a good explanation remains open. One intuitive property is that explanations should be sufficiently informative to reproduce the predictions given the data. In other words, a good explainer can be repurposed as a predictor. Post-hoc explainers do not achieve this goal as their explanations are highly dependent on fixed model parameters (e.g., learned GNN weights). To address this challenge, we propose RAGE (Robust Ante-hoc Graph Explainer), a novel and flexible ante-hoc explainer designed to discover explanations for graph neural networks using bilevel optimization, with a focus on the chemical domain. RAGE can effectively identify molecular substructures that contain the full information needed for prediction while enabling users to rank these explanations in terms of relevance. Our experiments on various molecular classification tasks show that RAGE explanations are better than existing post-hoc and ante-hoc approaches.
\end{abstract}

\section{Introduction}
\label{introduction}
A critical problem in machine learning on graphs is understanding predictions made by graph-based models in high-stakes applications. This has motivated the study of graph explainers, which aim to identify subgraphs that are both compact and correlated with model decisions. However, there is no consensus on what constitutes a good explanation. Recent papers \cite{ying2019gnnexplainer, luo2020parameterized, bajaj2021robust} have proposed different alternative notions of explainability that do not consider the user and instead are validated using examples. Other approaches have applied labeled explanations to learn an explainer directly from data \cite{faber2020contrastive}. However, datasets with such labeled explanations are hardly available.

\begin{figure}[htp] 
    \centering
    \includegraphics[keepaspectratio, width=0.8\textwidth]{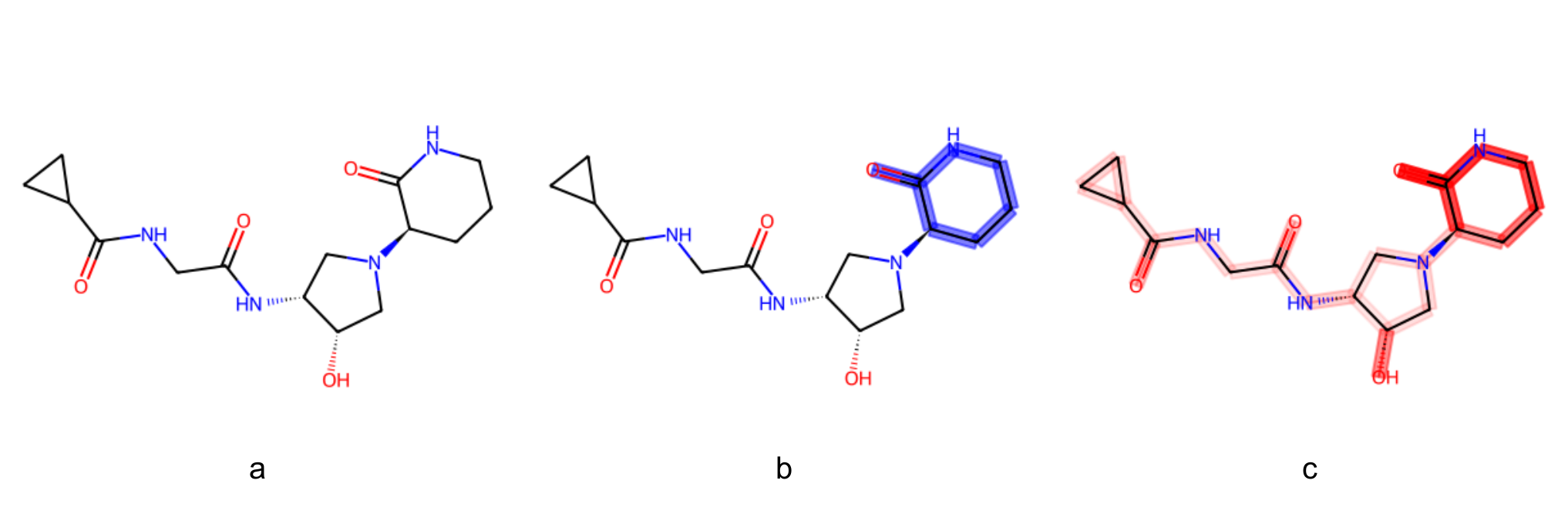}
    \caption{An example of explanations generated by our approach (RAGE) on molecular graphs from the synthetic dataset \textsc{LACTAM}. Subfigure (a) shows an input molecule, subfigure (b) highlights the ground-truth explanation and subfigure (c) illustrates the edge weights assigned by RAGE. Our method correctly identified the subgraph corresponding to the ground-truth lactam functional group.} %
    \label{fig::motivational_figure}
\end{figure}

\begin{figure}
\centering
\subfloat[Post-hoc Models]{
\includegraphics[keepaspectratio, width=0.42\textwidth]{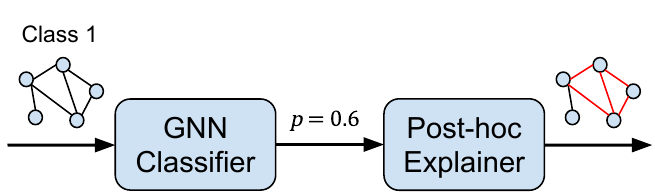}
}
\subfloat[Ante-hoc Models]{
\includegraphics[keepaspectratio, width=0.42\textwidth]{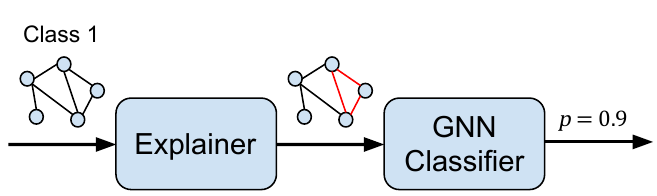}
}
\caption{(a) Post-hoc models generate explanations for a pre-trained GNN classifier using its predictions. (b) Ante-hoc models, as our approach, learn GNNs and explanations jointly. This enables ante-hoc models to identify GNNs that are both explainable and accurate. %
}
\label{fig::ante-post-hoc}
\end{figure}

Explainers can be divided into \textit{post-hoc} and \textit{ante-hoc} (or intrinsic) \cite{vilone2020explainable}. Post-hoc explainers treat the prediction model as a black box and learn explanations by modifying the input of a pre-trained model \cite{yuan2022explainability}. On the other hand, ante-hoc explainers learn explanations as part of the model. Figure \ref{fig::ante-post-hoc} compares the post-hoc and ante-hoc approaches in the context of graph classification. The key advantage of post-hoc explainers is their flexibility since they make no assumption about the prediction model to be explained or the training algorithm applied to learn the model. However, these explanations have two major limitations: (1) they are not sufficiently informative to enable the user to reproduce the behavior of the model, and (2) they are often based on a model that was trained without taking explainability into account.

The first limitation is based on the intuitive assumption that a good explanation should enable the user to approximately reproduce the decisions of the model for new input. That is simply because the predictions will often depend on parts of the input that are not part of the post-hoc explanations. The second limitation is based on the fact that for models with a large number of parameters, such as neural networks, there are likely multiple parameter settings that achieve similar values of the loss function. However, only some of such models might be explainable \cite{rudin2019stop,zhang2022protgnn}. While these limitations do not necessarily depend on a specific model, this paper addresses them in the context of Graph Neural Networks (GNNs) for graph-level classifications.

% This enables the user to select an appropriate trade-off between the compactness of the explanations and their discrimination power. 
This paper proposes RAGE, a novel ante-hoc explainer for graphs that finds compact explanations while maximizing the graph classification accuracy using bilevel optimization. RAGE explanations are given as input to the GNN, which guarantees that no information outside of the explanation is used for prediction. We show that RAGE explanations are more robust to noise in the input graph than existing (post-hoc and ante-hoc) alternatives. Moreover, our explanations are learned jointly with the GNN, which enables RAGE to learn GNNs that are accurate and explainable. In fact, we show that RAGE's explainability objective produces an inductive bias that often improves the accuracy of the learned GNN compared to the base model. We emphasize that while RAGE is an ante-hoc model, it is general enough to be applied to a broad class of GNNs.

% \textbf{\textcolor{red}{REWRITE THIS PARAGRAPH}}
% In our experiments, we quantitatively evaluate our RAGE on a variety of molecular benchmarks.
Figure \ref{fig::motivational_figure} shows an example of RAGE explanations on an input molecule from \textsc{Lactam}, one of our semi-synthetic molecular datasets. The goal is identifying whether the molecule contains lactam functional groups, which are cyclic amides of various ring sizes. Figure 1b shows the molecule with the ground-truth explanation highlighted in blue. Figure 1c illustrates the edge influences learned by RAGE, with darker shades of red indicating higher weights or more important edges. As expected, edges corresponding to the ground-truth explanation received distinguishably higher weights. We show more examples of explanations by RAGE in Appendix \ref{appendix-example}. Our main contributions can be summarized as follows: 

% Figure \ref{fig::motivational_figure} shows examples of RAGE explanations in two case studies. In \ref{fig::a}, we show an explanation from a synthetic dataset (\textsc{Planted Clique}), where the goal is to classify whether the graph has a planted clique or not based on examples. As expected, the edge influences learned by RAGE match with the planted clique. In \ref{fig::b}, we show an explanation for a real dataset (\textsc{Sunglasses}) with graphs representing headshots (images), where the goal is to classify whether the person in the corresponding headshot is wearing sunglasses. We notice that edge influences highlight pixels around the sunglasses. We provide a detailed case study using these two datasets in Section \ref{sec::case_study}, including a comparison against state-of-the-art post-hoc and ante-hoc explainers. We also evaluate our approach quantitatively in terms of accuracy, reproducibility, and robustness. Our results show that RAGE often outperforms several baselines. Our main contributions can be summarized as follows:

\begin{itemize}[leftmargin=*]
    
\item We highlight and empirically demonstrate two important limitations of post-hoc graph explainers. They do not provide enough information to enable reproducing the behavior of the predictor and are based on fixed models that might be accurate but not explainable.
    
\item We propose RAGE, a novel GNN and flexible explainer for graph classifications. RAGE applies bilevel optimization, learning GNNs in the inner problem and an edge influence function in the outer loop. Our approach is flexible enough to be applied to a broad class of GNNs.
    
\item We compare RAGE against state-of-the-art graph classification and GNN explainer baselines using 8 datasets---including 5 real-world ones. RAGE not only performs competitively or better compared to the baselines in terms of accuracy in most settings but also generates explanations that enable reproducing the behavior of the predictor. 

% We also provide additional case studies showing that our method improves interpretability by highlighting essential parts of the input data.
    
\end{itemize}

\section{Method}
\label{sec::methodology}

\subsection{Problem Formulation}

We formulate our problem as a supervised graph classification. A graph $G = (V,E)$ has node attributes $x_v$ for $v \in V$ and edge attributes $e_{uv}$ for $(u,v) \in E$. Given a graph set $\mathcal{G} = \{G_1, G_2, \dots, G_n\}$ and continuous or discrete labels $\mathcal{Y} = \{y_1, y_2, \dots, y_n\}$ for each graph respectively, our goal is to learn a function $\hat{f}:\mathcal{G}\to \mathcal{Y}$ that approximates the labels of unseen graphs. %

\begin{figure*}[b]
\centering
\includegraphics[keepaspectratio, width=\textwidth]{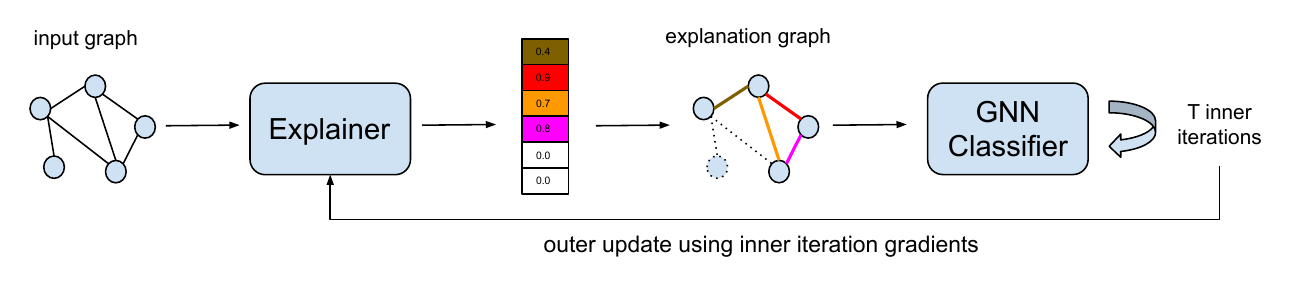}
\caption{Illustration of an edge-based ante-hoc explainer that uses bilevel optimization. Explainer generates an explanation graph from the input graph by assigning an influence value to each edge. Edge influences are incorporated to edge weights on the explanation graph, the input of GNN Classifier. The inner problem optimizes GNN Classifier with $T$ iterations, while the outer problem updates Explainer using gradients from inner iterations. The dotted edges in the explanation graph show that they do not influence the classification, while others have different degrees of influence.
\label{fig::ante_hoc_general_figure}}
\end{figure*}

\subsection{RAGE: Robust Ante-hoc Graph Explainer}

We introduce RAGE, an ante-hoc explainer that generates robust explanations using bilevel optimization. RAGE performs compact and discriminative subgraph learning as part of the GNN training that optimizes the prediction of class labels in graph classification tasks.

RAGE is based on a general scheme for an edge-based approach for learning ante-hoc explanations using bilevel optimization, as illustrated in Figure \ref{fig::ante_hoc_general_figure}. The explainer will assign an influence value to each edge, which will be incorporated into the original graph. The GNN classifier is trained with this new graph over $T$ inner iterations. Gradients from inner iterations are kept to update the explainer in the outer loop. The outer iterations minimize a loss function that induces explanations to be compact (sparse) and discriminative (accurate). We will now describe our approach (RAGE) in more detail. %

\subsubsection{Explainer - Subgraph Learning}

RAGE is an edge-based subgraph learner. It learns edge representations from edge features and the corresponding node representations/features. Surprisingly, most edge-based explainers for undirected graphs are not permutation invariant when calculating edge representations (e.g., PGExplainer \cite{luo2020parameterized} concatenates node representations based on their index order). Shuffling nodes could change their performance drastically since the edge representations would differ. We calculate permutation invariant edge representations $h_{ij}$ given two node representations $h_i$ and $h_j$ as follows: $h_{ij} = [\textbf{max}(h_i, h_j); \textbf{min}(h_i, h_j)]$, where $\textbf{max}$ and $\textbf{min}$ are pairwise for each dimension and $[\cdot;\cdot]$ is the concatenation operator. Additionally, most existing explainers do not consider edge features, which are crucial for certain domains such as molecular graphs. Since many recent GNNs readily incorporate edge features \cite{hu2020pretraining}, we utilize them in processing the input graph to obtain $h_i$ and $h_j$ that take into account both node and edge features.

Edge influences are learned via an MLP with sigmoid activation based on edge representations: $z_{ij} = MLP(h_{ij})$. This generates an edge influence matrix $Z\!\in\![0, 1]^{n \times n}$. We denote our explainer function as $g_{\Phi}$ with trainable parameters $\Phi$.

\subsubsection{Influence-weighted Graph Neural Networks}
\label{sec::extendedgnn}

Any GNN architecture can be made sensitive to edge influences $Z$ via a transformation of the adjacency matrix of the input graphs. As our model does not rely on a specific architecture, we will refer to it generically as $GNN(A, X)$, where $A$ and $X$ are the adjacency and attribute matrices, respectively. We rescale the adjacency matrix with edge influences $Z$ as follows: $A_Z = Z \odot A$.

The GNN treats $A_Z$ in the same way as the original matrix: $H =  GNN(A_Z, X)$

We generate a graph representation $h$ from the node representation matrix $H$ via a mean pooling operator. The graph representation $h$ is then given as input to a classifier that will predict graph labels $y$. Here, we use an MLP as our classifier.

\subsubsection{Bilevel Optimization}

In order to perform both GNN training and estimate the influence of edges jointly, we formulate graph classification as a bilevel optimization problem. In the inner problem (Equation \ref{eq:inner}), we learn the GNN parameters $\theta^* \in \mathbb{R}^h$ given edges influences $Z^* \in [0,1]^{nxn}$ based on a training loss $\ell^{tr}$ and training data $(D^{tr}, y^{tr})$. We use the symbol $C$ to refer to any GNN architecture. In the outer problem (Equation \ref{eq:outer}), we learn edge influences $Z^*$ by minimizing the loss $\ell^{sup}$ using support data $(D^{sup}, y^{sup})$. 

%The loss functions for the inner and outer problem, $f_{Z^*}$ and $F$, also apply regularization functions, $\Theta_{inner}$  and $\Theta_{outer}$, respectively.

% \begin{equation}
% \label{eq:outer}
%     Z^* = \argmin_Z F(\theta^*, Z) = \ell^{sup}(C(\theta^*, Z, D^{sup}), y^{sup}) + \Theta_{outer}
% \end{equation}
% \begin{equation}
% \label{eq:inner}
%     \theta^* = \argmin_\theta f_{Z^*}(\theta) = \ell^{tr}(C(\theta, Z^*, D^{tr}), y^{tr}) + \Theta_{inner}
% \end{equation}

\begin{equation}
\label{eq:outer}
    Z^* = \argmin_Z F(\theta^*, Z) = \ell^{sup}(C(\theta^*, Z, D^{sup}), y^{sup})
\end{equation}

\begin{equation}
\label{eq:inner}
    \theta^* = \argmin_\theta f_{Z^*}(\theta) = \ell^{tr}(C(\theta, Z^*, D^{tr}), y^{tr})
\end{equation}

RAGE can be understood through the lens of meta-learning. The outer problem performs \textit{meta-training} and edge influences are learned by a \textit{meta-learner} based on support data. The inner problem solves multiple \textit{tasks} representing different training splits sharing the same influence weights.

At this point, it is crucial to justify the use of bilevel optimization to compute edge influences $Z$. A simpler alternative would be computing influences as edge attention weights using standard gradient-based algorithms (i.e., single-level). However, we argue that bilevel optimization is a more robust approach to our problem. More specifically, we decouple the learning of edge influences from the GNN parameters and share the same edge influences in multiple training splits. Consequently, these influences are more likely to generalize to unseen data. We validate this hypothesis empirically using different datasets in our experiments in Appendix \ref{appendix-ablation}.

% \subsubsection{Loss Functions}

% RAGE loss functions have two main terms: a prediction loss and a regularization term. As prediction losses, we apply cross-entropy or mean-square-error, depending on whether the problem is classification or regression. The regularization for the inner problem $\Theta_{inner}$ is a standard $L_2$ penalty over the GNN weights $\theta$. For the outer problem $\Theta_{outer}$, we also apply an $L_1$ penalty to enforce the sparsity of $Z$. Finally, we also add an $L_2$ penalty on the weights of $g_{\Phi}$. %

\noindent

\subsection{Bilevel Optimization Training}
\label{sec:bilevel_opt_math}
% We further discuss the significance and the impact of this operation in Supp \ref{sec::no_reinit_GNN} \textbf{\textcolor{red}{CHECK}}.
% The main steps performed by our model (RAGE) are given in Algorithm \ref{algo:rage}. For each outer iteration (lines 1-14), we split the training data into two sets---training and support---(line 2). First, we use training data to calculate $Z^{tr}$, which is used for $GNN$ training in the inner loop (lines 5-10). Then, we apply the gradients from the inner problem to optimize the outer problem using support data (lines 11-13). Note that we reinitialize $GNN$ and $MLP$ parameters (line 4) before starting inner iterations to remove undesirable information \cite{zhou2022fortuitous} and improve data generalization \cite{alabdulmohsin2021impact}. We found reinitialization results in better performances in general. The main output of our algorithm is the explainer $g_{\Phi_{\kappa}}$. Moreover, the last trained $GNN_{\theta_{T}}$ can also be used for the classification of unseen data, or a new GNN can be trained based on $Z$. In both cases, the GNN will be trained with the same input graphs, which guarantees the behavior of the model $GNN_{\theta_{T}}$ can be reproduced using explanations from $g_{\Phi_{\kappa}}$.

The main steps performed by our model (RAGE) are given in Algorithm \ref{algo:rage}. For each outer iteration (lines 1-13), we split the training data into two sets---training and support---(line 2). First, we use training data to calculate $Z^{tr}$, which is used for $GNN$ training in the inner loop (lines 4-9). Then, we apply the gradients from the inner problem to optimize the outer problem using support data (lines 10-12). The main output of our algorithm is the explainer $g_{\Phi_{\kappa}}$. Moreover, the last trained $GNN_{\theta_{T}}$ can also be used for the classification of unseen data or a new GNN can be trained based on $Z$. In both cases, the GNN will be trained with the same input graphs, which guarantees the behavior of the model $GNN_{\theta_{T}}$ can be reproduced using explanations from $g_{\Phi_{\kappa}}$.

For gradient calculation, we follow the gradient-based approach described in \cite{grefenstette2019generalized}. The critical challenge of training our model is how to compute gradients of our outer objective with respect to edge influences $Z$. By the chain rule, such gradients depend on the gradient of the training loss with respect to $Z$. We will, again, use the connection between RAGE and meta-learning to describe the training algorithm.

\subsubsection{Training (Inner Loop)}

At inner loop iterations, we keep gradients while optimizing model parameters $\theta$.

\vspace{-0.1in}
$$
    \theta_{t+1} = inner\text{-}opt_t\left(\theta_{t}, \nabla_{\theta_t}\ell^{tr}(\theta_t, Z_{\tau})\right)
$$

After T iterations, we compute $\theta^*$, which is a function of $\theta_1, \dots, \theta_{T}$ and $Z_{\tau}$, where $\tau$ is the number of iterations for meta-training. Here, $inner\text{-}opt_t$ is the inner optimization process that updates $\theta_t$ at step $t$. If we use SGD as an optimizer, $inner\text{-}opt_t$ will be written as follows with a learning rate $\eta$:

\vspace{-0.1in}
$$
    inner\text{-}opt_t\left(\theta_{t}, \nabla_{\theta_t}\ell^{tr}(\theta_t, Z_{\tau})\right) \coloneqq \theta_{t} - \eta \cdot \nabla_{\theta_t}\ell^{tr}(\theta_t, Z_{\tau})
$$

\subsubsection{Meta-training (Outer Loop)}

After $T$ inner iterations, the gradient trajectory saved to $\theta^*$ will be used to optimize $\Phi$. We denote $outer\text{-}opt_{\tau}$ as outer optimization that updates $\Phi_{\tau}$ at step $\tau$. The meta-training step is written as:

\vspace{-0.1in}
\begin{align*}
    \Phi_{\tau+1} =& \ outer\text{-}opt_{\tau}\left(\Phi_{\tau}, \nabla_{\Phi_{\tau}}\ell^{sup}(\theta^*)\right) \\
    =& \ outer\text{-}opt_{\tau}\left(\Phi_{\tau}, \nabla_{\Phi_{\tau}}\ell^{sup}(inner\text{-}opt_T(\theta_{T}, \nabla_{\theta_T}\ell^{tr}(\theta_T, z_{\tau})))\right)
\end{align*}

\begin{algorithm}[t]
	\caption {RAGE}
	\label{algo:rage}
	\begin{algorithmic}[1]
	 \REQUIRE Graphs $A_{1:n}$, node attributes $X_{1:n}$, labels $y_{1:n}$, explainer $g_{\Phi_{0}}$, outer/inner loops $\kappa$ and $T$
	 \ENSURE Trained $g_{\Phi_\kappa}$
	 \FOR{$\tau \in [0, \kappa-1] $}
	 \STATE $A^{tr}, A^{sup}, X^{tr}, X^{sup}, y^{tr}, y^{sup} \leftarrow \text{split}(A_{1:n}, X_{1:n}, 
	 y_{1:n})$
	 \STATE $Z^{tr} \leftarrow g_{\Phi_{\tau}}(A^{tr}, X^{tr})$
	 % \STATE \text{(re)initialize} $GNN_{\theta_0}$, $MLP_{\theta_0}$
	 \FOR{$t \in [0, T-1]$}
	    \STATE $H^{tr} \leftarrow GNN_{\tau,t}(Z^{tr} \odot A^{tr}, X^{tr})$
	    \STATE $h^{tr} \leftarrow POOL_{mean}(H^{tr})$
	    \STATE $p^{tr} \leftarrow MLP_{\tau,t}(h^{tr})$
	    \STATE $GNN_{\tau,t+1}, MLP_{\tau,t+1} \leftarrow inner\text{-}opt \ \  f_{Z^{tr}}(p^{tr}, y^{tr})$
	 \ENDFOR
	 \STATE $Z^{sup} \leftarrow g_{\Phi_{\tau}}(A^{sup}, X^{sup})$
	 \STATE $p^{sup} \leftarrow MLP_{\tau,T}(POOL_{mean}(GNN_{\tau,T}(Z^{sup} \odot A^{sup}, X^{sup}))$
	 \STATE $g_{\Phi_{\tau + 1}} \leftarrow outer\text{-}opt \ \ F_{\theta_{T}}(p^{sup}, y^{sup})$
	 \ENDFOR
	 \RETURN{} $g_{\Phi_\kappa}$
	 \end{algorithmic}
\end{algorithm}

After each meta optimization step, we calculate edge influences $Z_{\tau+1}$ using $g_{\Phi_{\tau+1}}(.)$. Our training algorithm is more computationally intensive than training a simple GNN architecture. For that reason, we set $T$ to a small value. Note that both edge influence learning and graph classification use GNNs to produce node and graph representations. These processes can either use separate GNNs or share the same architecture, also known as weight sharing. We found the latter to produce slightly better accuracy. This choice makes sense because intuitively a good subgraph learner is inherently a good classifier. Such a strategy is also employed in \cite{miao2022interpretable}.

% \subsubsection{Using Shared Weights}
% Both edge influence learning and graph classification use GNNs to produce node and graph representations. These processes can either use separate GNNs or share the same architecture, also known as weight sharing. We found the latter to produce slightly better accuracy. This choice makes sense because intuitively a good subgraph learner is inherently a good classifier. Such a strategy is also employed in \cite{miao2022interpretable}.

% Therefore we set $T$ and $\kappa$ to small values. 

% Thus, RAGE can also be efficiently applied at training and testing time compared to competing baselines (Supp \ref{sec::running_time}). \textbf{\textcolor{red}{RUN THIS APPENDIX?}}

\section{Experiments}
\label{experiments}
 We evaluate RAGE on several datasets and compare it against both post-hoc and ante-hoc explainers in terms of discriminative power and robustness. A key advantage of our approach is being able to effectively search for a GNN that is both explainable and accurate. We focus our evaluation on molecular graphs, an important domain of graph learning with a wide range of critical scientific applications, such as drug discovery, which greatly benefit from good explanations.
 
\subsection{Datasets}
In our experiments, we use a total of 8 graph-based molecular datasets, including 5 real-world classification datasets and 3 synthetic molecular datasets with ground-truth labels that we curated.

\paragraph{Graph Classification Datasets} We consider 4 binary graph classification datasets from MoleculeNet \cite{wu2018moleculenet}: \textsc{BBBP}, \textsc{ClinTox}, \textsc{Tox21}, and \textsc{ToxCast}. Additionally, we include \textsc{Mutagenicity} \cite{kazius2005derivation,debnath1991structure,morris2020tudataset}, a popular molecular classification dataset with ground-truth explanations. However, since there is currently no consensus on these explanations \cite{tan2022learning,luo2020parameterized,debnath1991structure}, we exclude \textsc{Mutagenicity} from the evaluation of the explanations produced by RAGE and the baselines.

\paragraph{Constructing Molecular Synthetic Datasets} Most graph-based datasets with ground-truth explanations consist of synthetic non-molecular graphs \cite{agarwal2023evaluating}. Evaluating explainer models using these datasets may not reflect the expected behavior of these models in real-world settings (e.g., drug discovery) due to the disparity between synthetic graphs and real-world graphs. Instead, we curate 3 semi-synthetic molecular graph datasets with ground-truth explanations. We screen millions of molecules from the ChemBL database \cite{gaulton2012chembl} and extract the ones with either or both of Lactam and Benzoyl functional groups. Lactam groups are cyclic amides with various ring sizes and the Benzoyl group is a benzene ring attached to a carbonyl group. The 3 datasets are defined as follows:

\begin{itemize}[leftmargin=*]
    
\item \textbf{\textsc{Lactam}}: Positive if molecules containing a Lactam group and negative otherwise. There is no molecule with multiple Lactam groups.
    
\item \textbf{\textsc{BenLac}} (Benzoyl Lactam): Positive if molecules containing a Lactam group; negative if molecules containing a Benzoyl group, a Lactam group with a Benzoyl group, or no Lactam and Benzoyl groups. There is no molecule with multiple Lactam or Benzoyl groups.
    
\item \textbf{\textsc{BenLacM}} (Benzoyl Lactam Multiclass): Class I if molecules containing a Lactam group; class II if molecules containing a Benzoyl group. Each molecule has either a Lactam or a Benzoyl group. We do not term the labels positive or negative since there is no class of interest in this setting.
    
\end{itemize}

One favorable aspect of these datasets is that the prediction is simple for most classifiers yet the explanation is not trivial, as we show in Section \ref{classification-result} and Section \ref{explanation-result}. This property allows us to assess not only the quality of the explanation but also the ability of the classifier to pick up the right signals from data. Both aspects are important for RAGE and ante-hoc explainers as both the explanations and the classifier are learned together. More details on these datasets are provided in Appendix \ref{appendix-dataset-details}.

\subsection{Experimental Settings}
\label{experimental-settings}
\begin{table}
  \caption{Classification performances in AUC of RAGE and other baselines.}
  \label{benchmark}
  \centering
  \begin{adjustbox}{width=1\textwidth}
  \begin{tabular}{l|ccc|ccccc}
  \toprule
  Models     &  Lactam & BenLac & BenLacM &  BBBP & ClinTox & Tox21 & ToxCast & Mutagenicity \\
  \midrule
  RAGE (ours)    & 100.0 ± 0.0  & 99.6 ± 0.7 & 100.0 ± 0.0 & \underline{67.1 ± 2.1} & \underline{85.8 ± 4.2}  & \textbf{74.6 ± 0.7} & 62.1 ± 0.9 & \underline{87.7 ± 1.6}\\
  GSAT     & 100.0 ± 0.0 & 99.4 ± 1.6 & 100.0 ± 0.0 & 65.8 ± 2.2 & \textbf{87.5 ± 2.7} & 72.9 ± 0.9 & 61.4 ± 0.8 & 82.6 ± 1.3\\
  DIR-GNN  & 86.8 ± 4.7 & 89.7 ± 6.8 & 99.2 ± 0.9 & 64.4 ± 2.9 & 79.5 ± 6.0 & 70.8 ± 0.8 & 61.2 ± 0.6 & 84.3 ± 1.5\\
  \midrule
  GIN     & 100.0 ± 0.0 & 98.1 ± 3.9 & 100.0 ± 0.0 & 65.9 ± 1.9 & 83.5 ± 5.0 & \underline{74.3 ± 1.0} & 61.7 ± 0.7 & 87.4 ± 1.2\\
  GCN     & 100.0 ± 0.0 & 99.3 ± 1.0 & 100.0 ± 0.0 & 64.8 ± 2.8 & 83.4 ± 6.7 & 73.9 ± 0.6 & \underline{62.3 ± 1.0} & \textbf{88.0 ± 1.3}\\
  GAT     & 100.0 ± 0.0 & 98.3 ± 1.8 & 100.0 ± 0.0 & 65.1 ± 1.3 & 82.6 ± 4.5 & 74.0 ± 0.9 & \textbf{63.2 ± 0.8} & 87.5 ± 1.4\\
  GMT     & 100.0 ± 0.0 & 99.5 ± 0.6 & 100.0 ± 0.0 & 65.8 ± 1.9 & 82.3 ± 1.7 & 74.0 ± 0.8 & 62.0 ± 0.7 & 86.4 ± 1.1 \\
  VIB-GSL     & 88.2 ± 4.2 & 94.1 ± 4.3 & 99.6 ± 0.4 & \textbf{67.5 ± 1.9} & 78.8 ± 2.0 & 71.5 ± 4.9 & 61.1 ± 0.5 & 83.3 ± 1.4 \\
  \bottomrule
  \end{tabular}
  \end{adjustbox}
\end{table}

\paragraph{Baselines} We compare RAGE against a variety of baselines in terms of both classification and explanation abilities. For classification, we include classical models such as GCN \cite{kipf2016semi}, GAT \cite{velivckovic2017graph}, and GIN \cite{xu2018powerful}. We also include GMT \cite{baekkh21}, a state-of-the-art GNN with attention-based graph pooling, and VIB-GSL \cite{sun2022graph}, a method that applies graph structure learning. For explanations on the semi-synthetic molecular datasets, we compare RAGE against both inductive explainers (e.g., PGExplainer \cite{luo2020parameterized}, GEM \cite{lin2021generative}) and transductive explainers (e.g., GNNExplainer \cite{ying2019gnnexplainer}, CFF \cite{tan2022learning}). Most importantly, we compare RAGE against existing ante-hoc graph explainers such as GSAT \cite{miao2022interpretable} and DIR-GNN \cite{wu2021discovering}.

\paragraph{Models and Training}
For GSAT, we follow the configuration suggested by the authors for OGB datasets \cite{miao2022interpretable}. For DIR-GNN \cite{wu2021discovering} and VIB-GSL \cite{sun2022graph}, we try our best to finetune the models. For any baseline that does not support edge features, we replace their GNN implementation with ours. Specifically, for fair and consistent comparisons, we choose GIN \cite{xu2018powerful} with edge encodings as the backbone of all explainers. Following the experiments from previous works \cite{hu2020open} on molecular datasets, we construct the GIN backbone with 5 layers and 300 hidden dimensions. We also keep this architectural setting for other GNN baselines, including GCN \cite{kipf2016semi}, GAT \cite{velivckovic2017graph}, and GMT \cite{baekkh21}. For post-hoc baselines, we obtain their performances via GNN-X-Bench \cite{kosan2023gnnx}. For our experiments on RAGE, we use Adam optimizer, batch size 256, learning rate $1e-4$, and weight decay 0. All datasets are split with the 8:1:1 ratio for train, validation, and test splits. All experiments are conducted on a shared computing server with random access to either Nvidia V100 or  A100 GPUs.  

\subsection{Results on Graph Classification}
\label{classification-result}
Table \ref{benchmark} shows the results of RAGE and other baselines on chemical classification benchmarks. For each dataset, we report the mean AUC and error from 10 runs. The best performance on each dataset is bolded and the second best performance is underscored. We do not show these highlights for \textsc{Lactam}, \textsc{BenLac}, and \textsc{BenLacM} because most baselines obtain near-perfect classification on these semi-synthetic datasets. Nevertheless, these datasets are meant for comparing the explanation ability and perfect classification performances promote better comparison of the extracted explanations. We include them in Table \ref{benchmark} for the sake of completeness.

No single baseline obtains the best performance on more than one dataset. RAGE is the best classifier for \textsc{Tox21} while ante-hoc GSAT \cite{miao2022interpretable} performs the best on \textsc{ClinTox}. Substructure learner VIB-GSL \cite{sun2022graph} does particularly well on \textsc{BBBP} but not on other benchmarks. In terms of overall competitiveness, our method RAGE is among the top 2 performing models on 4 out of the 5 reported real-world chemical datasets. Notice that RAGE performs better than its backbone model GIN \cite{xu2018powerful} across all chemical benchmarks, confirming the effectiveness of our framework as a classifier. 

% Notice that while we use GIN \cite{xu2018powerful} as the backbone for RAGE in our experiment and in all benchmarks, RAGE performs better than the vanilla GIN. These results confirm the effectiveness of our framework as a classifier. 

\begin{table}
  \caption{Performance of RAGE and other baselines on explainability.}
  \label{explain-perf}
  \centering
  \begin{adjustbox}{width=1\textwidth}
  \begin{tabular}{l|cccccc}
    \toprule
    \multirow{2}{*}{Models}     & \multicolumn{2}{c}{Lactam}  &  \multicolumn{2}{c}{BenLac}  &  \multicolumn{2}{c}{BenLacM} \\
    \cmidrule(lr){2-3}
    \cmidrule(lr){4-5}
    \cmidrule(lr){6-7}
                & ExAUC  & Precision@10 & ExAUC & Precision@10 & ExAUC & Precision@10\\
    % Model     & SIDER   &  ClinTox & HIV   \\
    \midrule
RAGE (ours)  & \textbf{98.2 ± 1.6} & \underline{85.9 ± 6.8} & \underline{81.7 ± 7.9} & \textbf{87.0 ± 8.4} & \textbf{92.9 ± 2.8} & \textbf{83.6 ± 3.6} \\
GSAT & \underline{97.7 ± 1.9} & 80.1 ± 6.7 & \textbf{84.5 ± 8.3} & \underline{73.5 ± 10.6} & 74.1 ± 7.5 & 71.2 ± 11.5\\
% DIR-GNN &  &  &  &  &  &  \\
\midrule
PGExplainer & 96.9 ± 0.8 & \textbf{87.4 ± 4.1} & 73.3 ± 1.9 & 71.6 ± 2.6 & \underline{86.5 ± 1.8} & \underline{81.0 ± 0.1} \\
GNNExplainer & 80.1 ± 1.7 & 50.8 ± 3.4 & 51.7 ± 1.1 & 65.2 ± 6.8 & 57.6 ± 0.8 & 36.4 ± 1.5 \\
GEM & 68.1 ± 11.1 & 48.4 ± 17.9 & 55.0 ± 6.2 & 56.5 ± 8.9 & 56.6 ± 5.8 & 42.0 ± 5.5 \\
CFF & 93.5 ± 0.6 & 81.4 ± 1.3 & 50.7 ± 0.9 & 50.4 ± 2.3 & 51.4 ± 0.8 & 38.9 ± 1.2 \\
\bottomrule
\end{tabular}
\end{adjustbox}
\end{table}

\subsection{Results on Explainability}
\label{explanation-result}
In Table \ref{explain-perf}, we show the explanation performances of RAGE and other explainers on the 3 semi-synthetic molecular datasets. Two important metrics, explanation AUC and precision among the top 10 weighted edges, are reported based on the ground-truth subgraphs. We compare RAGE directly with GSAT \cite{miao2022interpretable}, a strong ante-hoc baseline. We also include post-hoc explainers of which explanations are based on the trained GIN models obtained from the graph classification experiment.  We use all graphs in calculating explanation AUCs and only use graphs with ground-truth explanations in calculating precisions. For more details about ground-truth explanations, refer to Appendix \ref{appendix-dataset-details}.

Across the datasets, RAGE performs competitively both in terms of explanation AUC and precision. Remarkably, on the multiclass dataset \textsc{BenLacM}, our method is the best in both metrics. On \textsc{Lactam}, RAGE obtains the best explanation AUC and the second best precision while on \textsc{BenLac}, we achieve the best precision and second best explanation AUC. The results suggest that ante-hoc methods like RAGE and GSAT generally do better than post-hoc alternatives, among which we find PGExplaner to be the most competitive.

% In \textsc{BenLac}, our precision is rather unstable and varies widely among different runs. This observation is likely due to the fact that \textsc{BenLac} has the most complicated labeling scheme among the semi-synthetic datasets by having misleading signals (refer to Appendix \ref{appendix-dataset-details} for more details). The other ante-hoc baseline GSAT has the best precision on \textsc{BenLac}, however, it performs poorly on \textsc{BenLacM}.

\begin{figure*}
\centering
\includegraphics[keepaspectratio, width=\textwidth]{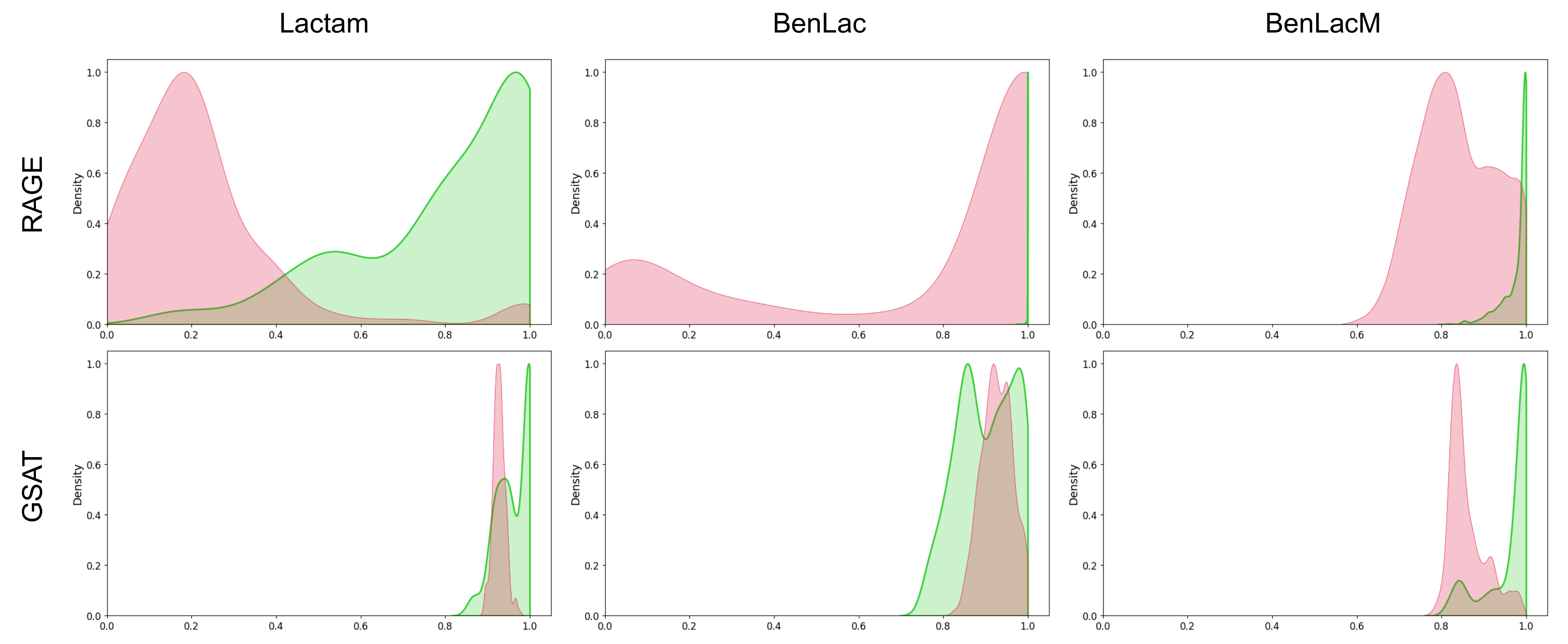}
\caption{Comparisons between the absolute weights assigned by RAGE and GSAT on ground-truth explanation (foreground) edges and the remaining (background) edges. The plots illustrate the density of edge weights across 3 synthetic molecular datasets. Densities of foreground edge weights are shown in green while those of background edge weights are shown in pink. The density of foreground edges by RAGE on \textsc{BenLac} is barely visible because most foreground weights are close to 1.0.}
\label{fig::rage_gsat_foreground_background}
\end{figure*}

Figure \ref{fig::rage_gsat_foreground_background} visualizes the edge weights assigned by RAGE and GSAT. The green density plots show weights assigned to edges belonging to the ground-truth explanations (foreground edges) while the pink density plots show the same for the remaining edges (background edges). Notice that since the semi-synthetic datasets are highly skewed, the actual density of foreground edges would be much smaller than that of background edges, however, for clarity, we normalize all plots. For GSAT, the weights assigned to either foreground edges and background edges are quite close to each other while RAGE tends to assign lower weights to background edges. In \textsc{BenLac}, RAGE assigns weights close to $1.0$ most foreground edges, which results in a density plot that looks like a vertical line. 

\subsection{Reproducibility}

\begin{figure*}
\centering
\includegraphics[keepaspectratio, width=\textwidth]{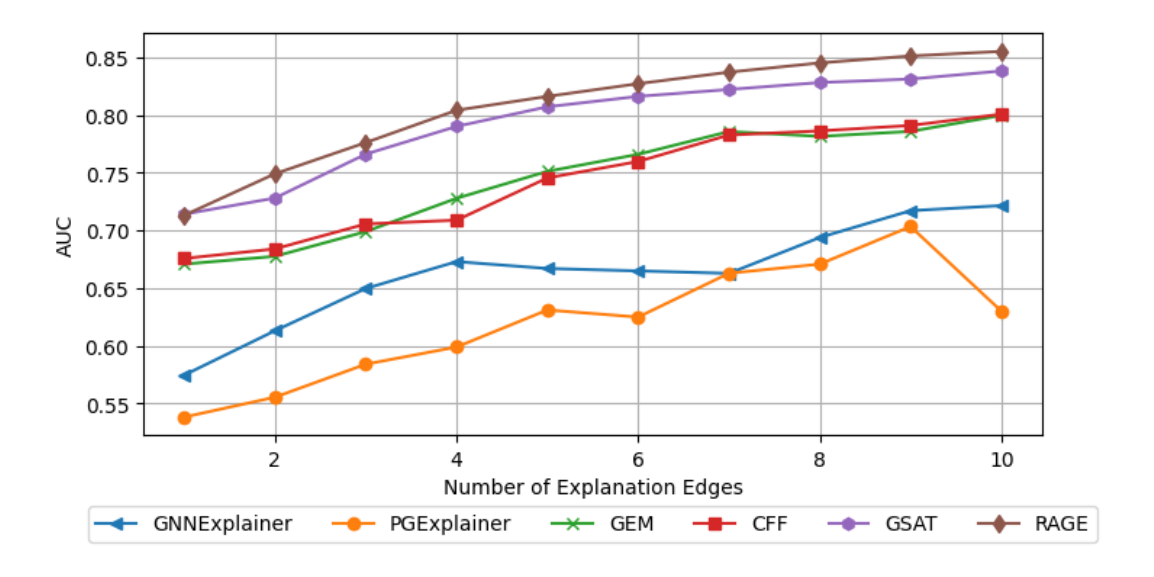}
\caption{Reproducibility of various ante-hoc and post-hoc explainers on Mutagenicity. We extracted explanation subgraphs by keeping the edges with the highest weights and trained a classifier on the resulting dataset comprising these subgraphs. RAGE and GSAT, being ante-hoc, perform better than other post-hoc explainers under such perturbation.}
\label{fig::reproducibility}
\end{figure*}

Reproducibility measures how explanations alone can predict class labels. It is a key property as it allows users to correlate explanations and predictions without neglecting potentially relevant information from the input. We vary the size of the explanations by thresholding edges based on their importance. We then train a GNN using only the explanations and labels. We compare RAGE against post-hoc and ante-hoc explainers on \textsc{Mutagenicity} and show the results in Figure \ref{fig::reproducibility}.

The results demonstrate that RAGE outperforms competing explainers in terms of reproducibility. Ante-hoc methods, RAGE and GSAT, emerge as the best and second best methods, confirming the superiority of ante-hoc learning in terms of generalization and robustness. RAGE incorporates these qualities through meta-training and bilevel optimization. Two post-hoc explainers, PGExplainer and GNNExplainer, perform poorly. As expected, larger explanations lead to better reproducibility.

\subsection{Ablation Study}
We include ablation experiments investigating the effects of varying the number of inner iterations and the backbone GNN on the performances of RAGE. The details can be found in Appendix \ref{appendix-ablation}.

\section{Related Works}
\label{sec::related_work}

% \textbf{\textcolor{red}{ADD GSAT, DIR, etc}}
% However, simple pooling operators that disregard the graph structure, such as mean and max, remain popular and have been shown to have comparable performance to more sophisticated alternatives \cite{mesquita2020rethinking}.
\textbf{Graph classification with GNNs:} Graph Neural Networks (GNNs) have gained prominence in graph classification due to their ability to learn features directly from data \cite{kipf2016semi,velivckovic2017graph,gilmer2017neural,xu2018powerful}. GNN-based graph classifiers aggregate node-level representations via pooling operators to represent the entire graph. The design of effective graph pooling operators is key for effective graph classification \cite{ying2018hierarchical, zhang2018end, li2019semi, gao2019graph, mesquita2020rethinking}. Recently, \cite{baekkh21} proposed a multi-head attention pooling layer to capture structural dependencies between nodes. In this paper, we focus on graph classification and show that our approach achieves competitive discriminative power compared to that of state-of-the-art alternatives \cite{xu2018powerful,baekkh21} while at the same time producing robust explanations.

% ProtGNN \cite{zhang2022protgnn} learns prototypes (interpretable subgraphs) for each class and makes predictions by matching input graphs and class prototypes. Bilevel optimization and prototypes help in generalizability of explanations. 
 % However, it is still outperformed by our ante-hoc explainer, which applies meta-training and bilevel optimization, in terms of different metrics. Our experiments show that RAGE explanations are more meaningful, faithful, and robust than alternatives and can reproduce the model behavior better than existing post-hoc and ante-hoc explainers.
\noindent
\textbf{Explainability of GNNs:} 
Explainability has become a key requirement for the application of machine learning in many settings (e.g., healthcare, court decisions, scientific discoveries) \cite{molnar2020interpretable}. Several post-hoc explainers have been proposed for explaining Graph Neural Networks' predictions using subgraphs \cite{ying2019gnnexplainer, luo2020parameterized, yuan2021explainability, bajaj2021robust, tan2022learning, xie2022task}. GNNExplainer \cite{ying2019gnnexplainer} applies a mean-field approximation to identify subgraphs that maximize the mutual information with GNN predictions. PGExplainer \cite{luo2020parameterized} applies a similar objective, but samples subgraphs using the \textit{reparametrization trick}. RCExplainer \cite{bajaj2021robust} generates counterfactual subgraph explanations. While post-hoc explainers treat a trained GNN as a black box ---i.e., it only relies on predictions made by the GNN---ante-hoc explainers are model-dependent. DIR-GNN trains an intrinsically explainable GNN by discovering and generating label-invariant rationales (subgraphs). GIB \cite{yu2021graph} applies the \textit{bottleneck principle} and bilevel optimization to learn subgraphs relevant for classification but different from the corresponding input graph. GSAT \cite{miao2022interpretable}, in an end-to-end fashion, combines information bottleneck with attention to sample explanation graphs as inputs into a classifier. This approach is the most related to our framework. We compare RAGE and GSAT rigorously in all experiments, in which RAGE outperforms the latter in various scenarios. However, the end-to-end training that GSAT employs is faster than our bilevel optimization. We discuss more about this limitation in Appendix \ref{appendix-limit}. Nevertheless, RAGE is competitive in terms of training and testing time compared to other GNN and explainer baselines (Appendix \ref{appendix-runtime}).

% Recently, TAGE \cite{xie2022task} proposes task-agnostic post-hoc graph explanations which also makes the explanations generalizable compared to existing post-hoc explainers.

\noindent
\textbf{Bilevel optimization:} 
Bilevel optimization is a class of optimization problems where two objective functions are nested within each other \cite{colson2007overview}. Although the problem is known to be NP-hard, recent algorithms have enabled the solution of large-scale problems in machine learning, such as automatic hyperparameter optimization and meta-learning \cite{franceschi2018bilevel}. Bilevel optimization has also been applied to graph problems, including graph meta-learning \cite{huang2020graph} and transductive graph sparsification \cite{wan2021edge}. 
% Like RAGE, GIB \cite{yu2021graph} also applies bilevel optimization to identify discriminative subgraphs inductively. However, we show that our approach consistently outperforms GIB in terms of discriminative power, reproducibility, and robustness.

\noindent
\textbf{Graph structure learning:} Graph structure learning (GSL) aims to enhance (e.g., complete, de-noise) graph information to improve the performance of downstream tasks \cite{zhu2021deep}. LDS-GNN \cite{franceschi2019learning} applies bilevel optimization to learn the graph structure that optimizes node classification. VIB-GSL \cite{sun2022graph} advances GIB \cite{yu2021graph} by applying a variational information bottleneck on the entire graph instead of only edges. We notice that GSL mainly focuses on learning the entire graph, whereas we only sparsify the graph, which reduces the search space and is more interpretable than possibly adding new edges. Furthermore, learning the entire graph is not scalable in large graph settings.

\section{Conclusion}

We investigated the problem of generating explanations for GNN-based graph-level classification and proposed RAGE, a novel ante-hoc GNN explainer based on bilevel optimization. RAGE inductively learns compact and accurate explanations by optimizing the GNN and explanations jointly. Moreover, different from several baselines, RAGE explanations do not omit any information used by the model, thus enabling the model behavior to be reproduced based on the explanations. We compared RAGE against state-of-the-art graph classification methods and GNN explainers using synthetic and real datasets. The results show that RAGE often outperforms the baselines on multiple evaluation metrics, including accuracy and robustness.
% \section*{References}

\begin{ack}
Mert Kosan worked on this project prior to joining Visa Inc.
\end{ack}

\bibliographystyle{plain}
\bibliography{main}

\newpage

\appendix

\section{Dataset Details}
\label{appendix-dataset-details}
\subsection{Graph Classification Benchmarks}
We provide more details about the datasets used in our study. Table \ref{data-stats} and Table \ref{synthetic-data-stats} show statistics regarding the classification benchmarks and our semi-synthetic datasets, respectively. All benchmarks are medium-size molecular graph datasets with both node features and edge features.

\begin{table}[!htbp]
    \caption{Molecular Classification Benchmarks}
    \label{data-stats}
    \centering
    \begin{adjustbox}{width=1\textwidth}
    \begin{tabular}{lcccc}
    \toprule
    Datasets & Number of Graphs & Number of Tasks & Number of Node Features & Number of Edge Features \\
    \midrule
    BBBP & 2039 & 1 & 9 & 3\\
    ClinTox & 1478 & 2 & 9 & 3\\
    Tox21 & 7831 & 12 & 9 & 3\\
    ToxCast & 8575 & 617 & 9 & 3\\
    Mutagenicity & 4337 & 1 & 14 & 3\\
\bottomrule
\end{tabular}
\end{adjustbox}
\end{table}

The molecular graph classification datasets reflect real-world applications of graph learning in chemistry. Most of them are obtained from the MoleculeNet \cite{wu2018moleculenet} with featurization following that of Open Graph Benchmark \cite{hu2020open}. In particular, there are 9 node featuers and 3 edge features describing various atom and bond properties. The Mutagenicity dataset is obtained from TUDataset \cite{morris2020tudataset}. In Mutagenicity, the node features and edge features are one-hot encodings of atom types and bond types. More specific descriptions of the datasets and the tasks are as follows:
\begin{itemize}
    \item BBBP: Blood-brain barrier permeability.
    \item ClinTox: Drugs that failed clinical trials for toxicity reasons
    \item Tox21: Toxicology on 12 biological targets
    \item ToxCast: Toxicology measurements via high-throughput screening
    \item Mutagenicity: Ability of an agent to cause genetic mutations
\end{itemize}

\begin{table}
  \caption{Semi-synthetic Molecular Benchmarks with Ground-truth Explanations.}
  \label{synthetic-data-stats}
  \centering
  \begin{adjustbox}{width=1\textwidth}
  \begin{tabular}{l|ccccc}
    \toprule
    Dataset & Task Type & Graph Type & Number of Graphs & Edge Label & Class \\
    \midrule
    \multirow{2}{*}{Lactam} &  \multirow{2}{*}{Binary}   & w lactam groups & 200 & Yes & Positive \\
                            &    & w/o lactam groups & 1000 & No & Negative \\
    \midrule
    \multirow{4}{*}{BenLac} &  \multirow{4}{*}{Binary}  & w lactam groups only & 100 & Yes & Positive \\
                            &    & w benzoyl groups only & 100 & Yes & Negative \\
                            &    & w both lactam and benzoyl groups & 200 & Yes & Negative \\
                            &    & w/o either lactam or benzoyl groups & 800 & No & Negative \\
    \midrule
    \multirow{2}{*}{BenLacM} &  \multirow{2}{*}{Multiclass}  & w lactam groups only & 500 & Yes & 1 \\
                            &    & w benzoyl groups only & 500 & Yes & 2 \\
\bottomrule
\end{tabular}
\end{adjustbox}
\end{table}

\subsection{Semi-synthetic Molecular Benckmarks}
We introduce 3 semi-synthetic molecular datasets with ground-truth labels (Table \ref{synthetic-data-stats}). Specifically, we screen millions of real molecules from ChemBL \cite{gaulton2012chembl} and randomly sample those with Lactam and/or Benzoyl substructures. Both of these functional groups are known to exhibit pharmaceutical values such as anti-bacterial effects. Additionally, lactam groups may have varying sizes, which is interesting as ground-truth explanations. The screening and processing are done via RDKit. Next, we describe the semi-synthetic datasets in more details. 

\textsc{Lactam} is a binary classification dataset to distinguish between molecules containing lactam groups and those not containing the lactam groups, with the former and the latter having the positive and negative labels, respectively. Benzoyl Lactam (\textsc{BenLac}) is also for binary classification but is a more challenging version of \textsc{Lactam}. In \textsc{BenLac}, we consider both the lactam and the benzoyl functional groups. Molecules may contain only lactam groups, only benzoyl groups, both lactam and benzoyl groups, or none of those groups. Out of these molecules, only those containing only lactam groups are the positive class. Benzoyl Lactam Multiclass \textsc{BenLacM} is a multiclass classification dataset. \textsc{BenLacM} requires distinguishing between molecules with either lactam groups or benzoyl groups. In this case, both classes have ground-truth explanations. Since most datasets on explainability consider binary classification with ground-truth explanations only for one class of interest, we believe it is interesting to look into scenarios in which different classes of data possesses different explanations. These datasets have the same node and edge featurizations as those of the MoleculeNet \cite{wu2018moleculenet} benchmarks (e.g \textsc{BBBP}).

\section{Ablation Study}
\label{appendix-ablation}

\subsection{Varying the Number of Inner Epochs and the Backbone GNN}
In Table \ref{ablation-layer-gnn}, we show the classification results in AUC of multiple RAGE trainings with different backbone GNNs and number of inner iterations. Except for \textsc{BBBP}, RAGE with multiple inner epochs tend to perform better than RAGE with only one inner epoch. Interestingly, more epochs do not guarantee equal or better performances. For example, RAGE with GIN backbone achieves the best AUC on \textsc{Tox21} using 5 inner epochs. The same observation holds for RAGE with GCN on \textsc{BBBP} and \textsc{ClinTox}. These observations suggest that the number of inner epochs is an important hyperparameter that should be carefully tuned in order to obtain the best performance when learning and explaning with RAGE. Since more inner epochs mean longer training time, depending on the use case, one might opt for less inner epochs, possible trading some diminishing performance gain for significantly better efficiency.

There is no clear winner between GIN and GCN as backbone GNNs. RAGE with GIN does better on \textsc{BBBP} and \textsc{ClinTox} while RAGE with GCN does better on \textsc{Tox21} and \textsc{ToxCast}. For that reason, it is important to experiment with multiple types of backbone graph learner for optimal results on any specific use case. 

\subsection{Learning With and Without Bilevel Optimization}
We create a single-level version of RAGE (RAGE-single) to test the effectiveness of our bilevel optimization scheme. More specifically, RAGE-single optimizes the explainer and GNN classifier parts in an end-to-end fashion with a single loss function. Table \ref{ablation-bilevel} shows that RAGE consistently outperforms RAGE-single. The performance gap is prominent across all chemical classification benchmarks. These observations confirm the effectiveness and necessity of meta-training and bilevel optimization in our framework.

\begin{table}[t]
  \caption{Ablation study on varying the GNN backbones and the number of inner epochs. The study is done on chemical classification benchmark with results shown in AUC.}
  \label{ablation-layer-gnn}
  \centering
  \begin{adjustbox}{}
  \begin{tabular}{lcccc}
  \toprule
  Models     &  BBBP & ClinTox & Tox21 & ToxCast \\
  \midrule
  $\text{GIN}_{1}$     & 66.8 ± 3.3 & 84.9 ± 4.3 & 72.8 ± 1.1 & 61.0 ± 1.5 \\
  $\text{GCN}_{1}$     & 66.5 ± 2.3 & 78.2 ± 5.2 & 74.1 ± 0.7 & 61.9 ± 2.7\\
  \midrule
  $\text{GIN}_{5}$     & 67.1 ± 2.1 & 85.8 ± 4.2  & 74.6 ± 0.7 & 62.1 ± 0.9 \\
  $\text{GCN}_{5}$     & 66.2 ± 1.8 & 85.2 ± 4.1 & 74.3 ± 0.6 & 62.5 ± 1.9 \\
  \midrule
  $\text{GIN}_{10}$     & 66.6 ± 2.4 & 85.7 ± 7.4 & 73.9 ± 1.0 & 62.0 ± 1.3\\
  $\text{GCN}_{10}$      & 65.2 ± 1.7 & 82.3 ± 9.2 & 74.6 ± 0.9 & 62.6 ± 1.4\\
  \bottomrule
  \end{tabular}
  \end{adjustbox}
\end{table}

\begin{table}[t]
  \caption{Ablation study on training with and without bilevel optimization. Classification results are shown in AUC. RAGE with bilevel optimization outperforms RAGE-single in every dataset. }
  \label{ablation-bilevel}
  \centering
  \begin{adjustbox}{}
  \begin{tabular}{lcccc}
  \toprule
  Models     &  BBBP & ClinTox & Tox21 & ToxCast \\
  \midrule
  RAGE (w bilevel optimization)     & 67.1 ± 2.1 & 85.8 ± 4.2  & 74.6 ± 0.7 & 62.1 ± 0.9 \\
  RAGE-single (w/o bilevel optimization)     & 60.0 ± 3.8 & 67.3 ± 6.0 & 65.3 ± 2.4 & 53.3 ± 2.1\\
  \bottomrule
  \end{tabular}
  \end{adjustbox}
\end{table}

\section{Running Time}
\label{appendix-runtime}
Bilevel optimization needs more training time than the standard GNNs. Table \ref{tab::running_times} shows training and testing times on the \textsc{Lactam} dataset for all methods. For training, we show, in seconds, the amount of time required to finish one training epoch while for testing, we show the amount of time taken to evaluate the whole test split. We do not include training time for GNNExplainer, GEM, and CFF since these are transductive models. Instead, we report, as testing time, the amount of time it take for these methods to fit and explain every graph in the testing set. 

RAGE trains slower than standard and some of the sophisticated GNN methods, while having comparable or faster testing time. For post-hoc graph explainers, RAGE is much faster in testing. For ante-hoc explainers, RAGE is slower than in training than GSAT, is the same in testing compared to GSAT, and is significantly faster than DIR-GNN in both training and testing.

\begin{table}[ht]
\caption{Training and testing time for RAGE and baselines for \textsc{Lactam}. }
\centering
\begin{tabular}{cccc}%
\toprule
& & Training (s/epoch) & Testing (s/fold) \\
\midrule
\multirow{3}{*}{Standard GNNs} & GCN & 0.165s & 0.021s \\
& GAT & 0.216s & 0.022s \\
& GIN & 0.203s & 0.021s \\
\midrule
\multirow{2}{*}{Sophisticated GNNs} & GMT & 0.268s & 0.027s \\
& VIB-GSL & 40.017s & 7.266s \\
\midrule
\multirow{4}{*}{Post-hoc Explainers} & PGExplainer & 78.026s & 1.212s \\
& GNNExplainer & N/A & 111.842s \\
& GEM & N/A & 88.401s \\
& CFF & N/A & 2306.452s \\
\midrule
\multirow{2}{*}{Ante-hoc Explainers} & GSAT & 0.331s & 0.025s \\
& DIR-GNN & 47.017s & 6.306s \\
\midrule
Our method & RAGE & 3.252s & 0.025s \\
\bottomrule
\end{tabular}

\label{tab::running_times}

\end{table}

\section{Limitations}
\label{appendix-limit}
The most critical limitation of RAGE is the training time (Table \ref{tab::running_times}), as we rely on bilevel optimization. Even though our inference time is quite efficient, long training time is still a problem in scenario such as online learning or lifelong learning. We expect this problem to be alleviated by applying more efficient bilevel optimization methods, such as those that apply stochastic samplings, decetralized processing, or single-loop algorithms \cite{yang2021provably,dagreou2022framework,dong2023single}. However, we did not experiment with different meta-learning or bilevel optimization approaches in our study and would like to leave this task for future work. 
% We also do not report training and inference time of RAGE and other baselines. Due to certain technical difficulties, we could not reliably reproduce consistent and comparable training time.

\section{More Examples of Explanations}
\label{appendix-example}
\begin{figure*}[b]
\centering
\includegraphics[keepaspectratio, width=0.9\textwidth]{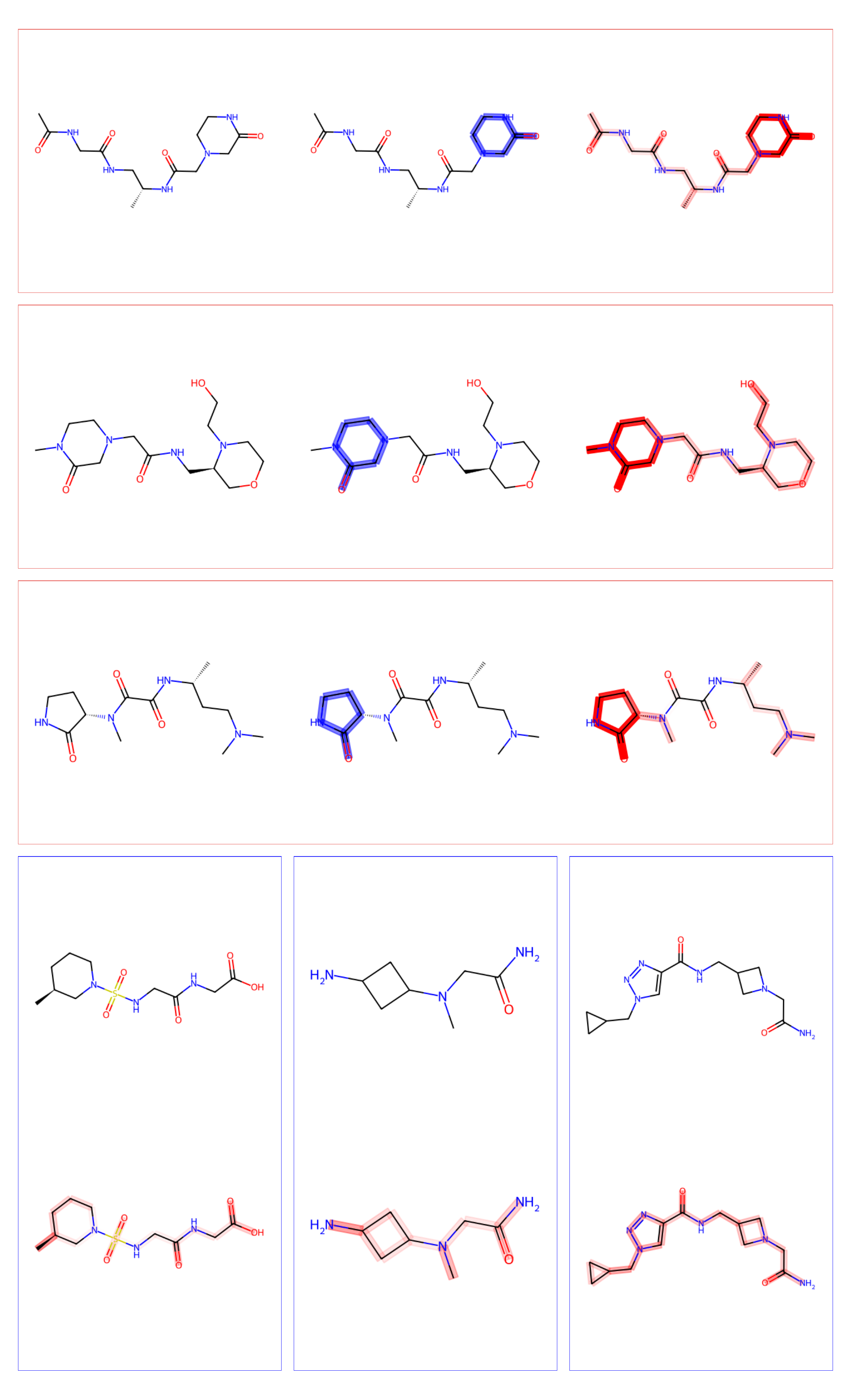}
\caption{Examples of explanations by RAGE on \textsc{Lactam}. Positive examples with ground-truth explanations are shown in red boxes and negative examples are shown in blue box.}
\label{fig::lactam_ex}
\end{figure*}

\begin{figure*}[b]
\centering
\includegraphics[keepaspectratio, width=0.9\textwidth]{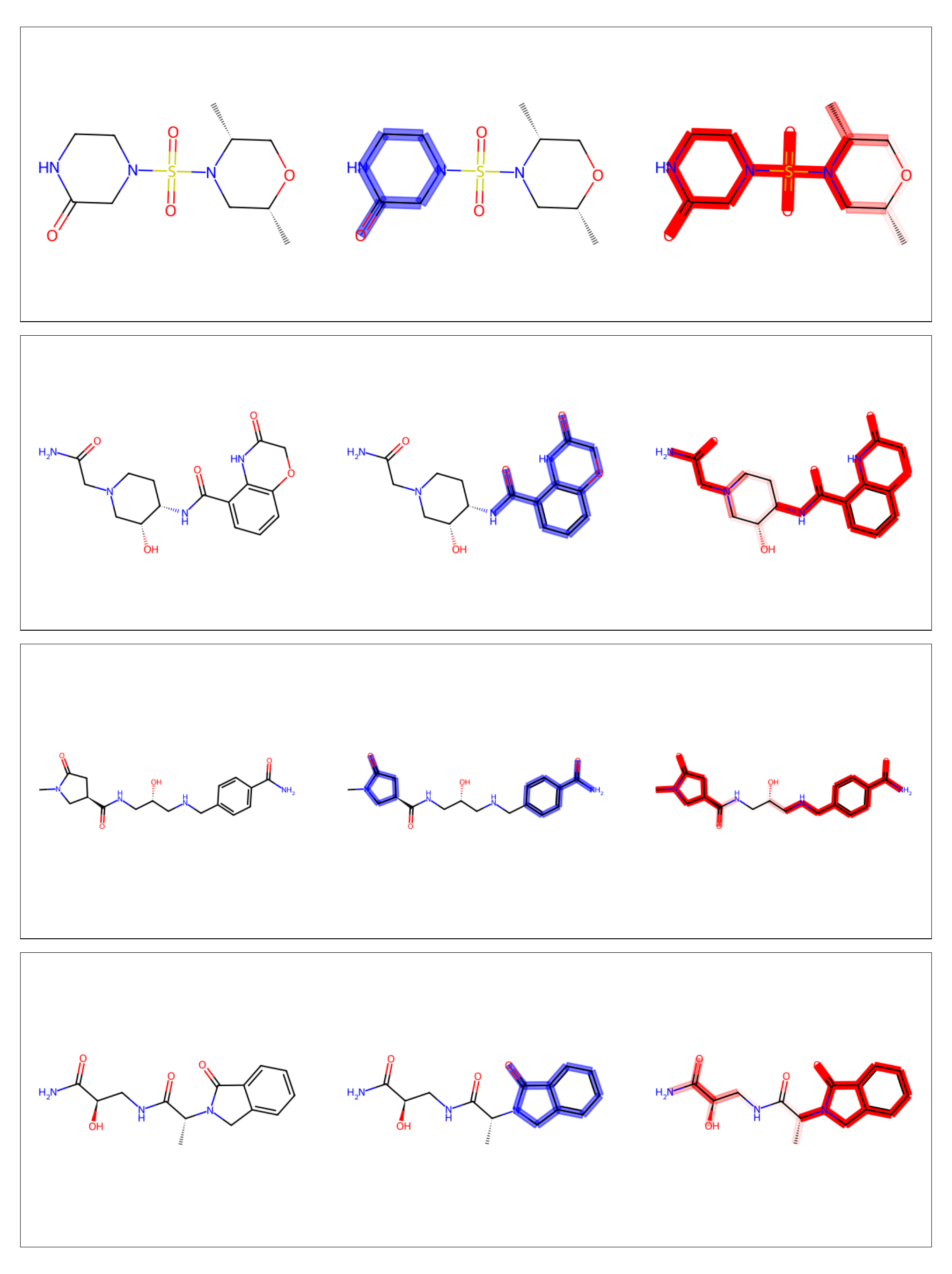}
\caption{Examples of explanations by RAGE on \textsc{BenLac}.}
\label{fig::benzoyl_lactam_ex}
\end{figure*}

\begin{figure*}[b]
\centering
\includegraphics[keepaspectratio, width=0.9\textwidth]{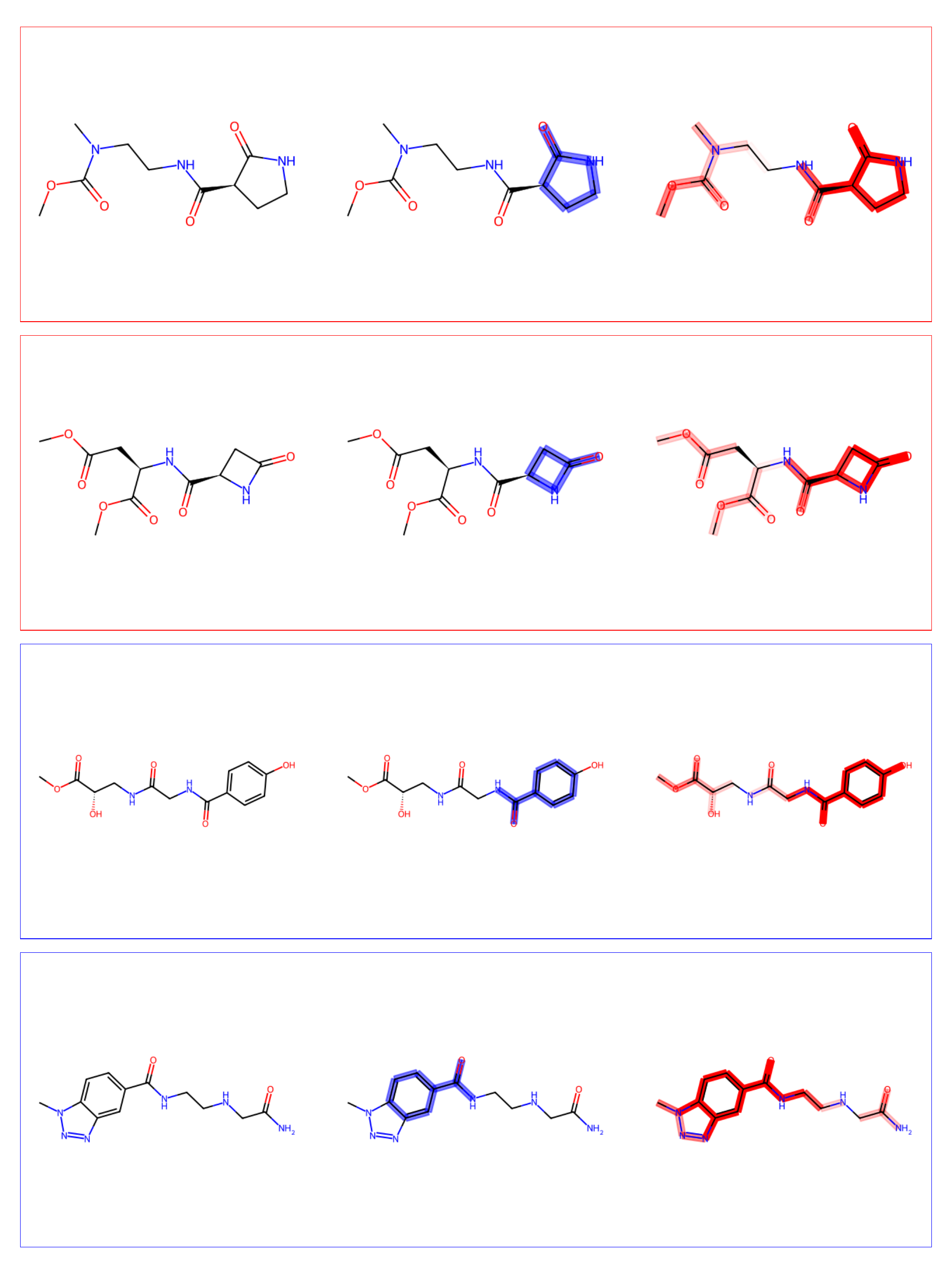}
\caption{Examples of explanations by RAGE on \textsc{BenLacM}. Red boxes and blue boxes distinguish examples from the 2 classes. Specifically for this dataset, we apply min-max scaling on the weights assigned by RAGE before plotting for better perceptibility.}
\label{fig::benzoyl_lactam_multiclass_ex}
\end{figure*}

We include more examples of the explanations discovered by RAGE. Figure \ref{fig::lactam_ex}, Figure \ref{fig::benzoyl_lactam_ex}, and Figure \ref{fig::benzoyl_lactam_multiclass_ex} show examples from \textsc{Lactam}, \textsc{BenLac}, and \textsc{BenLacM}, respectively. In these examples, blue highlights indicate ground-truth explanations and red highlights indicate edge importance learned by RAGE. Darker shades of red mean higher weights.

% Optionally include supplemental material (complete proofs, additional experiments and plots) in appendix.
% All such materials \textbf{SHOULD be included in the main submission.}

\end{document}